\documentclass{article}
\usepackage[utf8]{inputenc} 
\usepackage[T1]{fontenc}    
\usepackage{hyperref}       
\usepackage{url}            
\usepackage{booktabs}       
\usepackage{amsfonts}       
\usepackage{nicefrac}       
\usepackage{microtype}      
\usepackage{xcolor}         

\usepackage{graphicx}
\usepackage{amsmath}
\usepackage{booktabs}
\usepackage{caption}
\usepackage{authblk}
\usepackage{hyperref}
\usepackage{multirow}
\usepackage{subcaption}
\usepackage{natbib}
\usepackage{enumitem}
\usepackage{makecell} 
\usepackage{arydshln}

\usepackage[most]{tcolorbox}
\usepackage{etoolbox} 
\tcbset{
  colback=white,
  colframe=black!40,
  boxrule=0.5pt,
  arc=2mm
}
\newtcblisting{promptlisting}{
  breakable,
  listing only,
  listing options={
    basicstyle=\ttfamily\scriptsize,
    breaklines=true,
    showstringspaces=false,
    literate={
      {•}{{\textbullet}}1
      {’}{{\textquotesingle}}1
      {–}{{--}}1
      {-}{{---}}1
      {“}{{\textquotedblleft}}1
      {”}{{\textquotedblright}}1
    }
  }
}

\title{Can Continuous-Time Diffusion Models Generate and Solve Globally Constrained Discrete Problems? A Study on Sudoku}

\author{
Mariia Drozdova\\
University of Geneva\\
\texttt{mariia.drozdova@unige.ch}\\
\footnote{Code available at \url{https://github.com/MariiaDrozdova/sudoku_generation}}
}

\begin{document}

\maketitle

\begin{abstract}
Can standard continuous-time generative models represent distributions whose support is an extremely sparse, globally constrained discrete set? 
We study this question using completed Sudoku grids as a controlled testbed, treating them as a subset of a continuous relaxation space.
We train flow-matching and score-based models along a Gaussian probability path and compare deterministic (ODE) sampling, stochastic (SDE) sampling, and DDPM-style discretizations derived from the same continuous-time training.
Unconditionally, stochastic sampling substantially outperforms deterministic flows; score-based samplers are the most reliable among continuous-time methods, and DDPM-style ancestral sampling achieves the highest validity overall.
We further show that the same models can be repurposed for guided generation: by repeatedly sampling completions under clamped clues and stopping when constraints are satisfied, the model acts as a probabilistic Sudoku solver.
Although far less sample-efficient than classical solvers and discrete-geometry-aware diffusion methods, these experiments demonstrate that classic diffusion/flow formulations can assign non-zero probability mass to globally constrained combinatorial structures and can be used for constraint satisfaction via stochastic search.

\end{abstract}

\section{Introduction}
\label{sec:introduction}

Recent advances in generative modeling have demonstrated that complex data distributions can be learned by modeling the evolution of samples along continuous probability paths,  starting from a simple distribution such as an isotropic Gaussian and gradually transforming it into the data distribution. In this work, we closely follow the formulation of flow matching and score matching models introduced in \cite{holderrieth2026flow}\cite{holderrieth2025introductionflowmatchingdiffusion}.

We use Sudoku as a controlled stress test.
Valid completed grids form an extremely sparse subset, and “almost correct” samples are meaningless because validity requires simultaneously satisfying row, column, and box constraints.
We therefore study whether standard continuous-time flow and diffusion models can (i) generate valid Sudoku grids unconditionally and (ii) be repurposed for constraint satisfaction via guided sampling.
Our experiments compare deterministic versus stochastic samplers, flow-based versus score-based parameterizations, and continuous-time simulation versus DDPM-style discretizations derived from the same training objective.

We consider a Gaussian probability path that interpolates between a standard Gaussian and the data distribution.
For a data point $z\in\mathbb{R}^d$ we define
\begin{equation}
p_t(x\mid z)=\mathcal N(x;\alpha_t z,\beta_t^2 I),
\qquad
x_t=\alpha_t z+\beta_t\varepsilon,\ \ \varepsilon\sim\mathcal N(0,I),
\label{eq:path}
\end{equation}
with boundary conditions $\alpha_0=0,\beta_0=1$ and $\alpha_1=1,\beta_1=0$.

Flow matching learns a time-dependent vector field $u_\theta(x,t)$ whose ODE transports samples along the probability path.
Generation is performed by integrating the deterministic dynamics from noise at $t=0$ to $t=1$. This serves as a baseline for assessing whether deterministic transport alone can reach the valid Sudoku grids manifold.

Score-based models learn the score $\nabla_x\log p_t(x)$ along the same path and sample using stochastic differential equations.
Unlike deterministic flows, SDE sampling injects noise during generation, enabling exploration.
In our experiments this stochasticity is essential for obtaining valid Sudoku grids.
Further, the learned score can be fine-tuned and then evaluated on a discrete time grid to construct DDPM/DDIM-style samplers.

\paragraph{Contributions.}
This work provides an empirical study of continuous-time generative models on a highly
constrained discrete problem. Our main contributions are:
\begin{itemize}
    \item We present a comparison of flow matching, score matching, and
    discrete diffusion models (DDPM/DDIM) on Sudoku, isolating the effects of deterministic
    versus stochastic sampling, continuous-time versus discrete-time dynamics, and flow-
    versus score-based parameterizations.
    
    \item For \emph{unconditional generation}, we show that stochastic sampling is essential
    and that discrete diffusion models (DDPM/DDIM) substantially outperform continuous-time
    SDE and ODE samplers.
    
    \item For \emph{guided Sudoku solving}, we show that the relative performance landscape
    changes: linear probability paths combined with score-based SDEs and $\beta(t)$-scaled
    noise perform competitively, despite deviating from probability-path–consistent
    diffusion dynamics.
    
    \item We provide a detailed empirical analysis of sampling stability and runtime behavior,
    highlighting failure modes and the role of stochasticity (e.g.\ diffusion noise in SDEs
    and inference-time dropout) in avoiding collapse.
\end{itemize}

\section{Related Literature}

Generative models are widely used in high-fidelity image generation \cite{ho2020denoising,song2020score,karras2022elucidating}. 
More recently, there has been growing interest in using iterative generative dynamics as a computational mechanism for reasoning- and constraint-heavy tasks, including emerging directions in language modeling \cite{jolicoeurmartineau2025morerecursivereasoningtiny}. 

Our work follows the standard continuous-time formulation of flow- and score-based models as presented in \cite{holderrieth2026flow}. 
We also use trigonometric parameterizations of probability paths that unify diffusion and flow matching from \cite{lu2024simplifying}.

A particularly relevant early success on Sudoku is the Dirichlet Diffusion Score Model (DDSM) \cite{avdeyev2023dirichlet}, which introduces a principled way to apply continuous-time diffusion to discrete variables by diffusing in the probability simplex with a Dirichlet stationary distribution.
This representation is well suited to categorical variables, forcing samples to remain on the simplex throughout sampling.
With sufficient sampling, DDSM can solve all puzzles in a hard 17-clue benchmark via conditional generation, albeit with a very large sampling budget on difficult instances.
In contrast, we deliberately avoid such structured representations and instead study whether standard diffusion models, trained purely via distribution matching and allowing intermediate invalid states, can learn globally constrained solutions from data alone.

Complementary to representation-driven diffusion approaches, Iterative Reasoning through Energy Diffusion (IRED) \cite{du2024learning} frames reasoning as explicit optimization over learned, annealed energy landscapes \cite{lecun2006tutorial}.
Rather than sampling solutions via a reverse diffusion process, IRED learns an energy function whose minima correspond to valid solutions and performs iterative refinement at inference time.
In contrast to this explicit energy-minimization framework, our work adheres to the standard diffusion and flow formulation and empirically studies its behavior on globally constrained discrete problems without introducing explicit energy functions.

\section{Methods}
\label{sec:methods}

\subsection{Dataset}
We train on the same train/test split of sudoku dataset as in \cite{jolicoeurmartineau2025morerecursivereasoningtiny}. Each grid is represented as a $9\times 9$ array with entries in $\{1,\dots,9\}$, and is flattened to a length-$81$ sequence of cells. For our continuous generative models, each cell is encoded as a $9$-dimensional vector of logits, yielding a tensor of shape $(81,9)$ per grid.

\subsection{Probability path}
We use a simple linear Gaussian probability path, defined by
\[
x_t = \alpha_t z + \beta_t \varepsilon,
\qquad
\alpha_t = t,\quad \beta_t = 1 - t,
\]
where $z$ denotes a clean Sudoku grid encoded in $\mathbb{R}^{81\times 9}$, $\varepsilon\sim\mathcal N(0,I)$, and $t\in[0,1]$. This choice satisfies the boundary conditions $\alpha_0=0$, $\beta_0=1$ and $\alpha_1=1$, $\beta_1=0$, and corresponds to a linear interpolation between pure noise and the data distribution.

However, this linear path does not correspond to a diffusion forward process in the sense of
denoising diffusion probabilistic models (DDPMs), since it does not satisfy
$\alpha(t)^2 + \beta(t)^2 = 1$. Thus, for diffusion we use a different parametrization:

\[
x_t = \alpha_t z + \beta_t \varepsilon,
\qquad \alpha(t)=\sin\!\Big(\frac{\pi}{2}t\Big),
\qquad
\beta(t)=\cos\!\Big(\frac{\pi}{2}t\Big),
\]

More details are provided in Appendix~\ref{app:ddpm}.

\subsection{Neural Parameterization}
\label{subsec:arch}

To parameterize the time-dependent vector field (or score function), we use a lightweight Transformer operating on the $81$ Sudoku cells as tokens (one token per cell).
Each token carries a $9$-dimensional state (logits), so the input is $z\in\mathbb{R}^{81\times 9}$ together with a scalar time $t\in[0,1]$, and the model outputs an update of the same shape.

Each per-cell state is first projected into a hidden embedding of dimension $H=128$.
To reflect Sudoku structure, we add learned positional embeddings for row, column, and $3\times 3$ box indices.
Time conditioning is implemented using a standard Fourier feature embedding of dimension $64$, followed by a small MLP that maps the time embedding into the hidden dimension and is added to every token.

The backbone consists of $L=4$ Transformer blocks with multi-head self-attention ($8$ heads) and position-wise MLPs, using residual connections, LayerNorm, and dropout ($0.01$).
The final hidden states are normalized and projected back to $\mathbb{R}^{9}$ per cell.
Across experiments, flow and score networks share the same architecture (approximately $3.3$M parameters) and differ only in the training objective.
Additional architectural details are provided in Appendix~\ref{app:model}.

 Both flow and score models were trained separately for $300{,}000$ iterations using a batch size of $2000$ on an NVIDIA RTX~3090 GPU. All the checkpoints are released in the GitHub repository.

\subsection{Training Objectives}

\paragraph{Velocity parameterization.}
For the Gaussian probability path $x_t = \alpha_t z + \beta_t \varepsilon$, the target velocity field admits a closed-form expression
\[
u_t^{\text{target}}(x \mid z)
=
\Big(\dot{\alpha}_t - \frac{\dot{\beta}_t}{\beta_t}\alpha_t\Big) z
\;+\;
\frac{\dot{\beta}_t}{\beta_t}\, x ,
\]
where dots denote derivatives with respect to $t$. In the flow-matching setting, we train a neural network $u_\theta(x,t)$ to directly approximate this target velocity field by minimizing a mean squared error objective between $u_\theta(x_t,t)$ and $u_t^{\text{target}}(x_t \mid z)$, with $(x_t,z)$ sampled from the Gaussian path.

\paragraph{Score-based parameterization.}
The corresponding conditional score along the same path is given analytically by
\[
\nabla_x \log p_t(x \mid z) = -\frac{x-\alpha_t z}{\beta_t^2}.
\]
Directly learning this expression was found to be numerically unstable in our setting, particularly near $t\to 1$ where $\beta_t\to 0$ and the score magnitude diverges. Instead, we parameterize a rescaled score model by training a neural network to predict the bounded quantity
\[
\beta_t^2 \nabla_x \log p_t(x \mid z) = -(x-\alpha_t z).
\]
We denote this network output by $\tilde{s}_\theta(x,t)$.
This rescaling avoids explicit division by $\beta_t^2$ and significantly improves numerical stability of the training.

\paragraph{Connection between flow matching and score matching objectives.}
For Gaussian paths, the velocity field and the score function are analytically related. As a result, one may train either component and derive the other explicitly. In our experiments, we consider both approaches and study their implications for sampling and stability.

\[
s_\theta(x,t)
=
\frac{\alpha(t)\,u_\theta(x,t) - \dot\alpha(t)\,x}
{\beta(t)^2\,\dot\alpha(t) - \alpha(t)\,\dot\beta(t)\,\beta(t)}.
\]

\[
u_\theta(x,t)
=
\frac{\dot{\alpha}(t)}{\alpha(t)}\,x
+
\frac{\beta(t)^2 \dot{\alpha}(t)
      - \alpha(t)\beta(t)\dot{\beta}(t)}
     {\alpha(t)}\,
s_\theta(x,t).
\]

Once the score (or/and the velocity field) has been learned, sampling proceeds by numerically simulating the corresponding stochastic differential equation using Euler-Maruyama.

\paragraph{Time sampling during training.}
For satisfying global Sudoku constraints late-time behavior is important, thus, we choose training times non-uniformly. Specifically, we draw $u\sim\mathcal U(0,1)$ and set
\[
t = 1 - (1-u)^2,
\]
which biases samples toward larger values of $t$. In addition, we clamp the time variable away from one by enforcing $t \le 1 - 10^{-4}$ to avoid numerical issues. This was enough to avoid training collapse for flow-based training, however for score-based training we have to additionally perform scaling as described above.

\subsection{Sampling procedures}
\label{subsec:sampling}

We generate samples by numerically simulating either deterministic or stochastic dynamics
defined by the learned vector field and/or score.
Let $\{t_k\}_{k=0}^K$ denote a discretization of $[0,1]$ with step size $h_k=t_{k+1}-t_k$.

\paragraph{Deterministic flow sampling.}
As a baseline, we simulate the learned flow field
\[
\frac{dx_t}{dt} = u_\theta(x_t,t)
\]
using forward Euler updates,
\[
x_{t_{k+1}} = x_{t_k} + h_k\,u_\theta(x_{t_k},t_k).
\]
This corresponds to standard flow-matching generation and produces deterministic trajectories.

\paragraph{Stochastic sampling via SDEs.}
To introduce stochasticity, we consider SDEs of the form
\[
dX_t =
\Big(
u_\theta(X_t,t)
+
\frac{1}{2}\sigma^2 s_\theta(X_t,t)
\Big)dt
+
\sigma\,dW_t,
\]
where $s_\theta(x,t)\approx\nabla_x\log p_t(x)$ is a learned score function and $\sigma>0$
controls the noise magnitude.
When the score is exact, this construction preserves the target probability path $p_t$
\cite{holderrieth2026flow}.
We discretize this SDE using the Euler--Maruyama method.

\paragraph{Empirical $\beta(t)$-scaled noise.}
In addition to constant-$\sigma$ diffusion, we also consider an empirical variant in which
the stochastic term is scaled by the path variance $\beta(t)$.
Concretely, the Gaussian noise term in the Euler--Maruyama update is multiplied by $\beta(t_k)$.

\[
dX_t =
\Big(
u_\theta(X_t,t)
+
\frac{1}{2}\sigma^2 s_\theta(X_t,t)
\Big)dt
+
\beta(t)\,dW_t,
\]

Although this modification no longer corresponds to an exact probability-path SDE,
we find that it substantially improves numerical stability and success rates on Sudoku.

We emphasize that this sampler should be interpreted as a guided stochastic search process
rather than an exact sampler of $p_t$.
All experimental results explicitly distinguish between constant-$\sigma$ and $\beta(t)$-based
sampling.

\paragraph{DDPM-style discretization.}
Finally, although models are trained in continuous time, we also evaluate a DDPM-style
ancestral sampler obtained by discretizing the same Gaussian probability path.
This procedure converts the learned score into a noise predictor and directly inverts the
corresponding Gaussian noising process.
Details of this discretization and its derivation from the continuous path are provided in
Appendix~\ref{app:ddpm_from_path}.

\subsection{Sudoku solving}
\label{subsec:guided_sampling}

Sudoku solving corresponds to sampling \emph{guided} completions given a set of fixed digits.  
Let $z_{\mathrm{given}}\in\mathbb R^{81\times 9}$ denote the sudoku input (partially filled sudoku where zero correspond to unknown), and let
$m\in\{0,1\}^{81\times 1}$ be the corresponding mask indicating constrained cells:
\[
m_i = 1 \ \text{if cell $i$ is given},\qquad m_i=0 \ \text{otherwise}.
\]
We write $\odot$ for elementwise multiplication with broadcasting over the digit dimension.

\paragraph{Hard clamping (inpainting).}
The simplest constraint mechanism clamps givens after each stochastic step:
\[
\Pi_{\mathrm{hard}}(x) \;=\; m\odot z_{\mathrm{given}} + (1-m)\odot x.
\]

\paragraph{Two-stage soft $\rightarrow$ hard constraints (path-consistent injection).}
Hard clamping at early (high-noise) times can be too restrictive because it forces clean digits into a highly noisy state.  
To address this, we use a two-stage rule controlled by a threshold $\tau\in[0,1]$ :

\begin{itemize}
\item \textbf{Soft (path-consistent) injection} for $t\le \tau$: we replace givens by a noisy version consistent with the forward path,
\[
x^{\mathrm{given}}_t = \alpha(t)\,z_{\mathrm{given}} + \beta(t)\,\varepsilon,\qquad \varepsilon\sim\mathcal N(0,I),
\]
and set
\[
\Pi_{\mathrm{soft}}(x,t) \;=\; m\odot x^{\mathrm{given}}_t + (1-m)\odot x.
\]
\item \textbf{Hard clamping} for $t>\tau$: we enforce exact givens via $\Pi_{\mathrm{hard}}$.
\end{itemize}

\subsection{Evaluation metrics}

\paragraph{Validity rate.}
For unconditional generation, we report the fraction of generated samples that correspond
to fully valid Sudoku grids. A grid is considered valid only if it satisfies all Sudoku
constraints exactly (rows, columns, and $3\times 3$ subgrids). Partial or nearly-valid
grids are not counted.

\paragraph{Entropy.}
To measure model confidence during sampling, we compute the mean per-cell entropy of the
relaxed output distribution,
\[
H = \frac{1}{81}\sum_{i=1}^{81} \Big(-\sum_{k=1}^9 p_{i,k}\log p_{i,k}\Big),
\]
where $p_{i,k}$ denotes the predicted probability of digit $k$ in cell $i$. Entropy
provides a local measure of uncertainty, with low entropy indicating near one-hot
predictions. We also visualize entropy trajectories over time to study convergence
behavior.

\paragraph{Constraint violations.}
After discretizing the relaxed output via a per-cell argmax, we count the number of
violations of Sudoku constraints. Specifically, we measure the number of duplicate digits
in each row, column, and $3\times 3$ box, reporting their average. This metric captures the degree of global
constraint satisfaction even when full validity is not achieved.

\section{Results}
\label{sec:results}

\begin{figure}[t]
    \centering
    \includegraphics[width=0.95\linewidth]{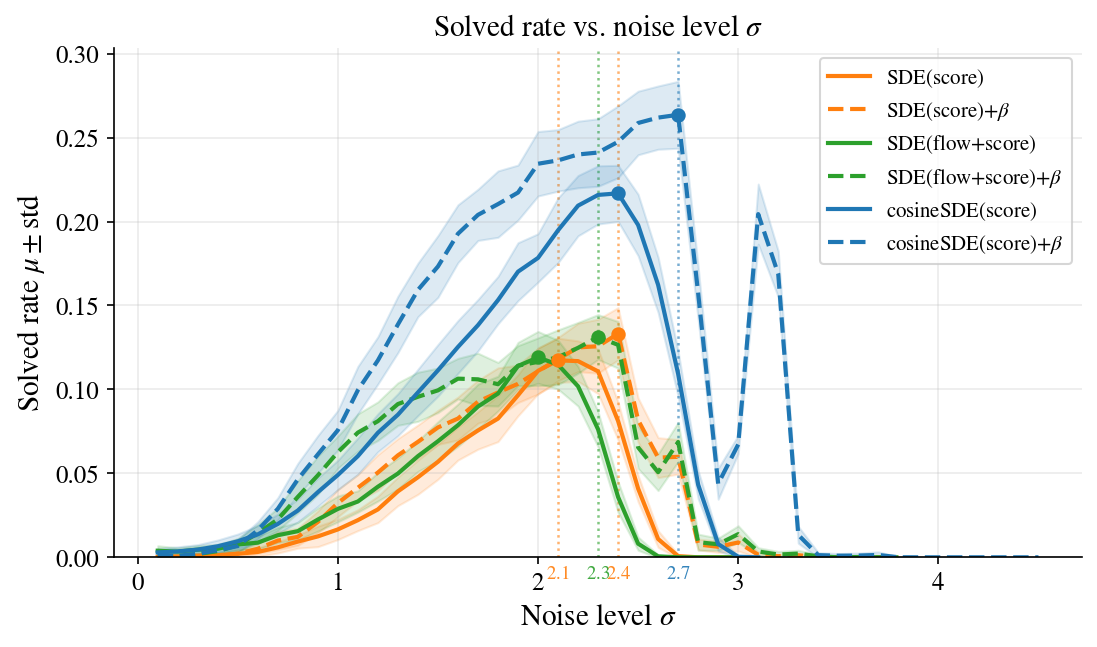}
    \caption{
        Comparison of the solved rate metric as a function of the noise level~$\sigma$
        for SDE with only a trained score model(flow derived analytically from it), and SDE with both score and flow trained; both either with diffusion coefficient $\sigma$ or $\beta(t)$.
        Markers indicate the maximum value for each curve, with the corresponding
        $\sigma$ annotated.
    }
    \label{fig:sde_first_number_vs_sigma}
\end{figure}

\subsection{Unconditional Sudoku Generation}

\paragraph{Evaluation protocol.}
For unconditional generation, we generate batches of $500$ samples and repeat each experiment
$50$ times with independent noise realizations.
A generated grid is counted as valid only if it satisfies all Sudoku constraints
(rows, columns, and $3\times 3$ subgrids).
We report the mean success rate across runs together with its standard deviation.

\paragraph{Compared sampling strategies.}
We train flow models and score models independently and evaluate all sampling procedures
that can be constructed from them.
From a trained \emph{flow} model, we consider deterministic ODE sampling and stochastic SDE sampling,
where the score term is reconstructed analytically from the learned flow.
From a trained \emph{score} model, we analogously consider the induced ODE, standard SDE sampling,
and SDE sampling with path-dependent noise scaling.
Finally, we evaluate SDEs constructed from \emph{both} a learned flow and a learned score.
These configurations isolate the effects of deterministic versus stochastic sampling,
flow-based versus score-based parameterizations,
and the choice of diffusion coefficient in the simulator.
For the cosine schedule we additionally study sampling with DDPM and DDIM using the score model.

\paragraph{Effect of stochasticity and noise level for linear probability path.}
Figure~\ref{fig:sde_first_number_vs_sigma} shows the solved rate as a function of the noise scale
$\sigma$ for different SDE constructions under the linear probability path.
Across all configurations, performance exhibits a pronounced dependence on $\sigma$,
with a clear optimal region for each method.
Too little noise typically leads to premature collapse into invalid configurations,
while excessive noise prevents convergence to discrete Sudoku solutions.

Across score-based and flow-based SDEs, $\beta(t)$-scaled noise shifts the optimal region
toward higher effective noise levels and increases the peak success rate.

SDEs derived purely from a learned flow model perform poorly across all noise levels;
we therefore omit them from Figure~\ref{fig:sde_first_number_vs_sigma}
and report their results only in Table~\ref{tab:linear_results}.

Table~\ref{tab:linear_results} summarizes unconditional generation results under the linear
Gaussian probability path. Deterministic ODE sampling achieves near-zero success rates across all variants, indicating that deterministic transport alone
is insufficient to reach the highly constrained Sudoku solution manifold and that stochastic sampling is essential in this setting. Stochastic samplers are reported using the optimal noise level found in Figure~\ref{fig:sde_first_number_vs_sigma}.

\begin{table}[ht]
\centering
\caption{Linear schedule for probability path coefficients $\alpha(t)$ and $\beta(t)$}
\label{tab:linear_results}

\scriptsize
\setlength{\tabcolsep}{-0.8pt}
\renewcommand{\arraystretch}{1.15}

\begin{tabular}{l *{8}{>{\centering\arraybackslash}p{1.35cm}}}
\hline
\textbf{Metric} &
\rotatebox{80}{SDE+Score+$\beta$} &
\rotatebox{80}{SDE+Score} &
\rotatebox{80}{SDE+Score+Flow+$\beta$} &
\rotatebox{80}{SDE+Score+Flow} &
\rotatebox{80}{SDE+Flow+$\beta$} &
\rotatebox{80}{SDE+Flow} &
\rotatebox{80}{ODE+Score} &
\rotatebox{80}{ODE+Flow} \\
\hline
\hline
Rate (50 runs) & 0.1332 & 0.1173 & 0.1312 & 0.1191 & 0.0041 & 0.0048 & 0.0003 & 0.0035 \\
Std. of rate & 0.0155 & 0.0132 & 0.0134 & 0.0155 & 0.00248 & 0.00294  & 0.00069 & 0.003113 \\
\hline
Pearson corr & -0.88 & -0.30 & -0.91 & -0.29 & -0.23 & -0.29 & -0.26 & -0.27 \\
$p$-value & $\approx10^{-166}$ & $\approx10^{-11}$ & $\approx10^{-190}$ & $\approx10^{-10}$ & $\approx10^{-7}$ & $\approx10^{-10}$ & $\approx10^{-9}$ & $\approx10^{-9}$\\
Steps & 200 & 200 & 200 & 200 & 200 & 200 & 200 & 200 \\
Optimal $\sigma$ & 2.4 & 2.1 & 2.3 & 2.0 & 0.4 & 0.3 & - & - \\
\hline
\end{tabular}
\end{table}

SDEs constructed from learned score models achieve success rates of approximately
$12\%$-$13\%$, substantially outperforming both deterministic ODE sampling and
SDEs derived purely from learned flow models (rightmost columns), $0.3\%$-$0.4\%$.
This suggests that learning the score function is more efficient when
sampling from highly constrained discrete distributions.

Combining a learned score with a learned flow does not lead to further gains in this
setting: the joint flow+score SDE performs comparably to the score-only SDE at their
respective optimal noise levels.
Across stochastic variants, using the path-dependent diffusion coefficient $\beta(t)$ improves both mean performance and run-to-run stability.
The strongest results under the linear schedule are obtained by
\texttt{SDE+Score+$\beta$} and \texttt{SDE+Score+Flow+$\beta$}.

As a diagnostic, we also report the Pearson correlation between entropy and the number
of constraint violations.
Entropy is averaged over a late-time window (integration steps $30$ to $15$ before
termination), and constraint violations are computed on the corresponding discretized
outputs.
Across methods, a strong negative correlation indicates that lower entropy is predictive
of fewer violations, suggesting that entropy provides a useful proxy for convergence
quality and the model’s uncertainty about global constraint satisfaction.

\begin{figure}[t]
\centering
\setlength{\tabcolsep}{6pt}

\begin{tabular}{ccc}
\includegraphics[width=0.30\textwidth]{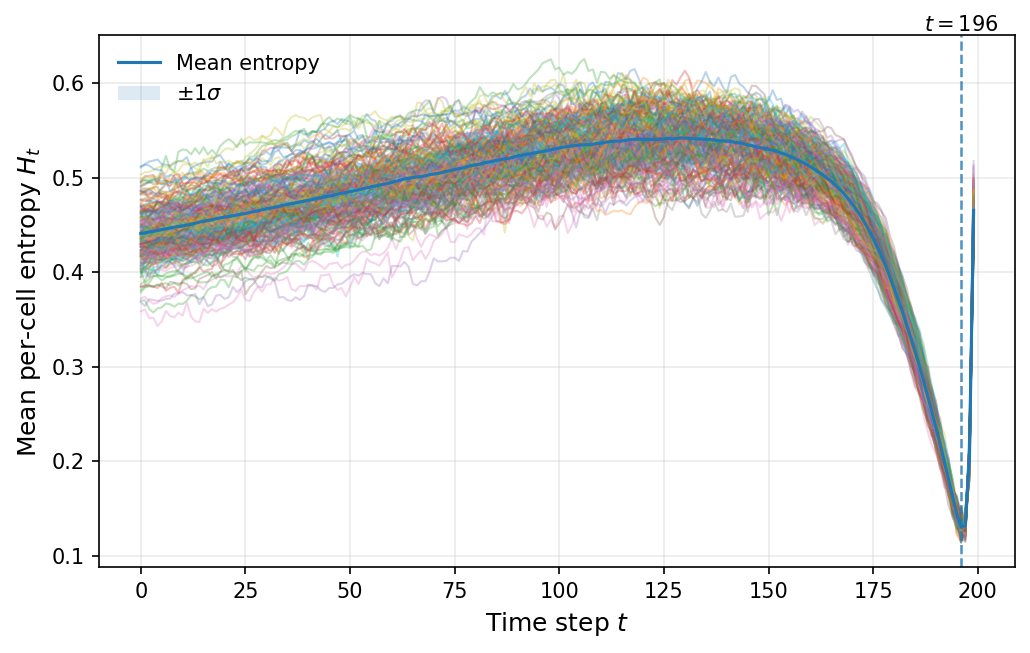} &
\includegraphics[width=0.30\textwidth]{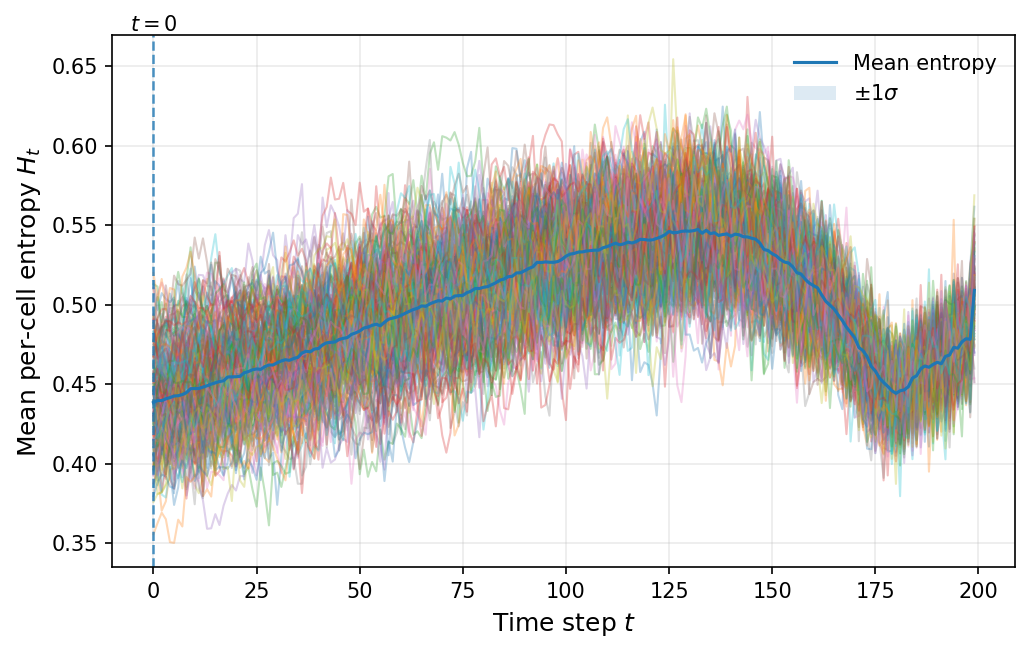} &
\includegraphics[width=0.30\textwidth]{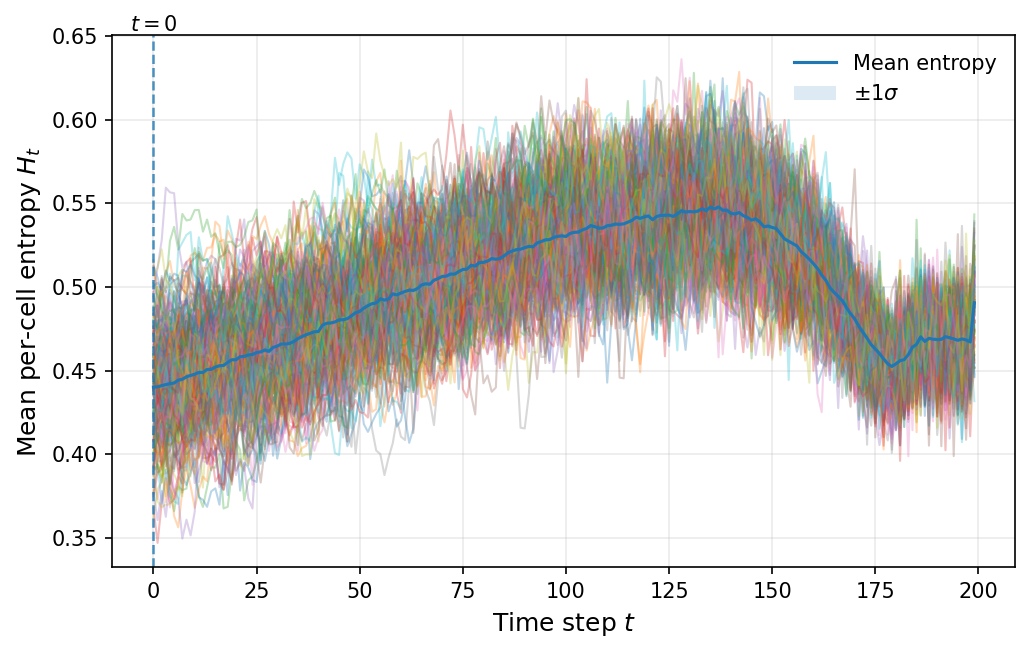}\\

\small SDE (flow) &
\small SDE (flow+score) &
\small SDE (score)\\[6pt]

\includegraphics[width=0.30\textwidth]{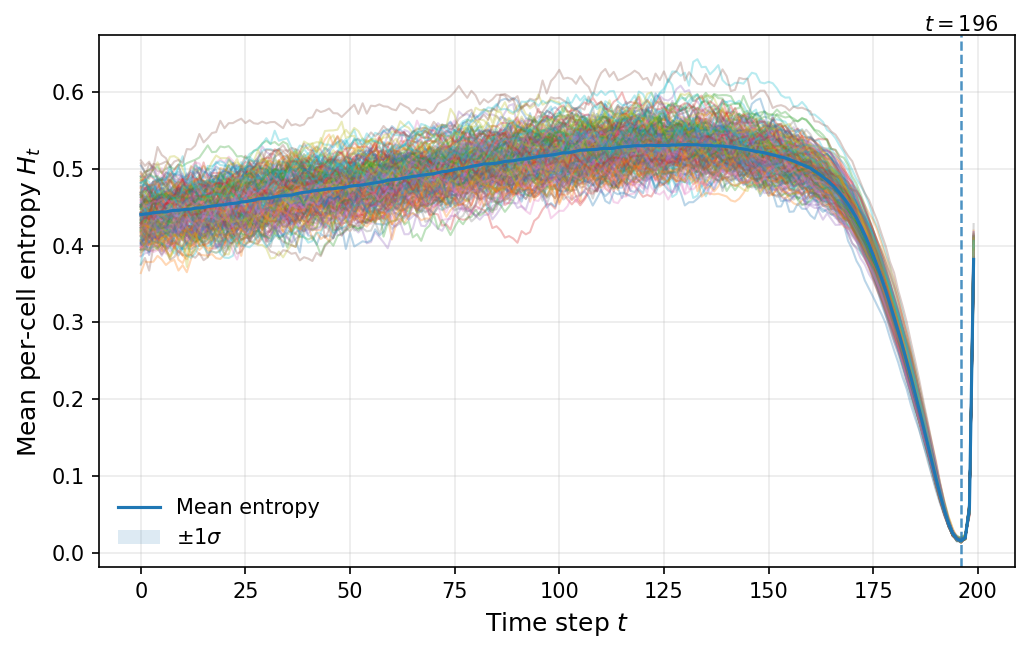} &
\includegraphics[width=0.30\textwidth]{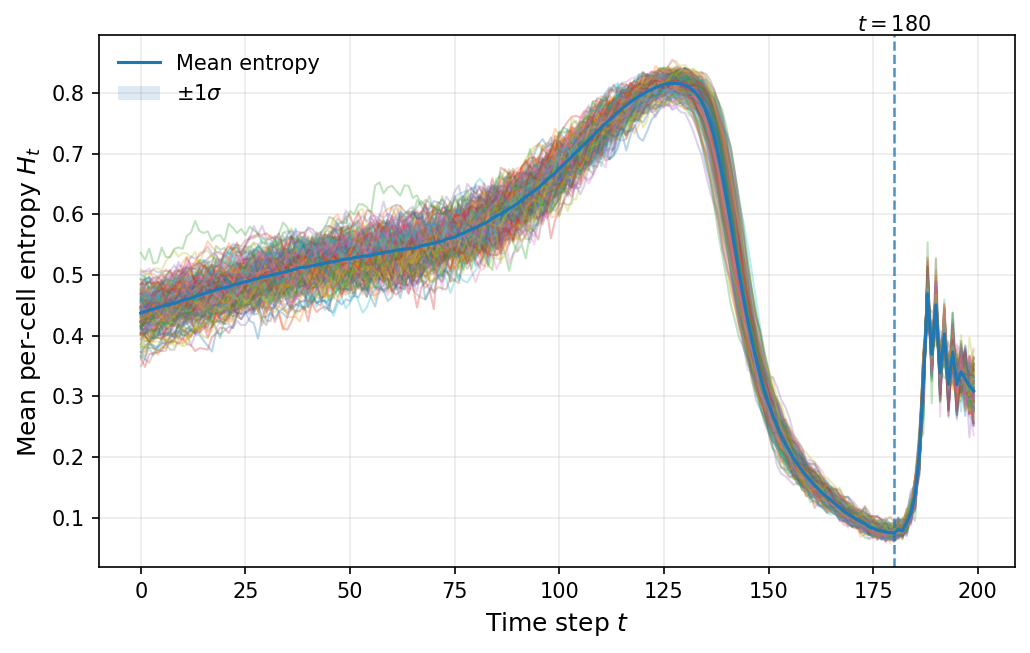} &
\includegraphics[width=0.30\textwidth]{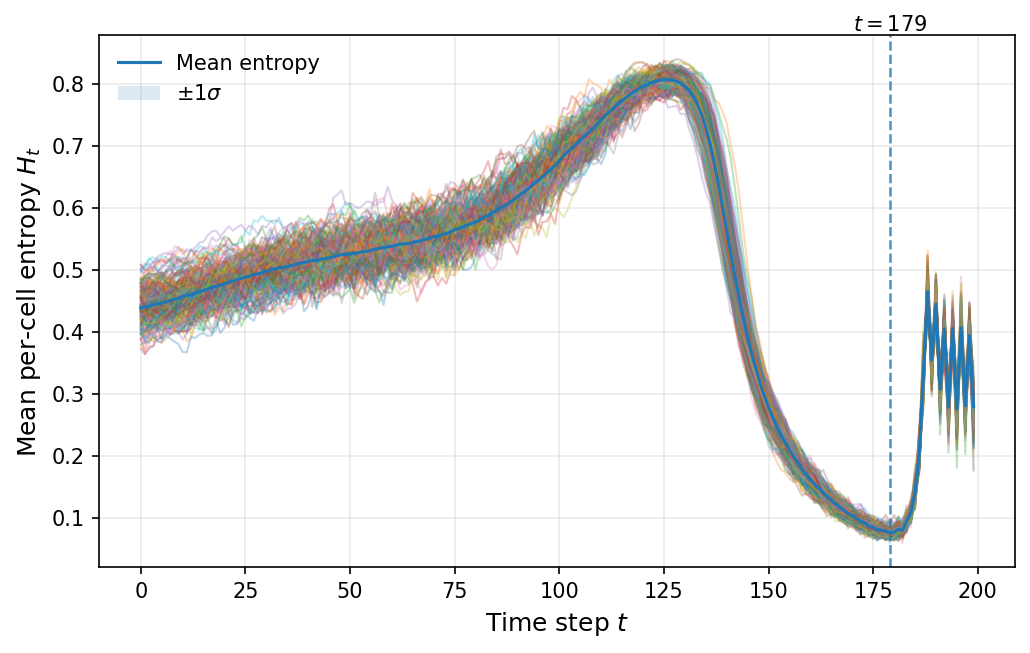}\\

\small SDE (flow+$\beta$) &
\small SDE (flow+score+$\beta$) &
\small SDE (score+$\beta$)
\end{tabular}

\caption{
Entropy under linear schedule. For each plot we chose $\sigma$ which had the best success rate.
Top row shows SDE variants with canonical Brownian noise;
bottom row shows the corresponding variants with $\beta_t$ instead of $\sigma$ in the simulator.
Columns compare having only score model trained or only flow trained, or combining both flow and score models trained separately.
}
\label{fig:entropy_linear_sde}
\end{figure}

Figure~\ref{fig:entropy_linear_sde} shows the evolution of mean per-cell entropy for
linear-schedule SDE samplers evaluated at their respective optimal noise levels.
Across all stochastic variants, we observe a consistent qualitative pattern:
entropy initially increases as noise dominates, then decreases as global structure
emerges, followed by a small late-stage increase.

Samplers using the path-dependent diffusion coefficient $\beta(t)$ typically reach
a lower minimum entropy than their constant-$\sigma$ counterparts.
This behavior reflects the different noise profiles induced by the two choices.
With a constant diffusion coefficient, large noise levels are applied uniformly
throughout inference, whereas $\beta(t)$-scaled diffusion naturally anneals the noise
as $t\to1$, reducing stochasticity during the final refinement stage.
Near convergence, $\beta(t)$-based samplers also exhibit entropy increase and
oscillations, which may arise from the fact that the empirical $\beta(t)$ sampler no
longer exactly follows the prescribed probability path.

Despite this deviation, $\beta(t)$-based samplers consistently achieve higher final
success rates.
This indicates that strict adherence to the probability path is not required for
effective convergence, and that annealed stochasticity can be beneficial for solving
globally constrained discrete problems.

\begin{table}[ht]
\centering
\caption{Cosine schedule for probability path coefficients $\alpha(t)$ and $\beta(t)$ and diffusion discretizations.}
\label{tab:cosine_results}

\scriptsize
\setlength{\tabcolsep}{3pt}
\renewcommand{\arraystretch}{1.15}

\begin{tabular}{l *{5}{>{\centering\arraybackslash}p{1.35cm}}}
\hline
\textbf{Metric} &
\rotatebox{60}{SDE+$\beta$} &
\rotatebox{60}{SDE} &
\rotatebox{60}{DDIM} &
\rotatebox{60}{DDPM} &
\rotatebox{60}{ODE} \\
\hline
\hline
Rate (50 runs) & 0.2636 & 0.2168 & 0.7880 & 0.8364 & 0.0015 \\
Std. of rate & 0.0199 & 0.0167 & 0.0147 & 0.0138 & 0.0009 \\
\hline
Pearson corr & -0.65 & -0.61 & -0.89 & -0.15 & -0.14 \\
$p$-value & $\approx 10^{-61}$ & $\approx 10^{-53}$ & $\approx 10^{-3}$ & $\approx 10^{-174}$ & $\approx 10^{-3}$ \\
Steps & 200 & 200 & 400 & 400 & 200 \\
Optimal $\sigma$ & 2.7 & 2.4 & -- & -- & -- \\
\hline
\end{tabular}
\end{table}

\begin{figure}[t]
\centering
\setlength{\tabcolsep}{6pt}

\begin{tabular}{ccc}
\includegraphics[width=0.30\textwidth]{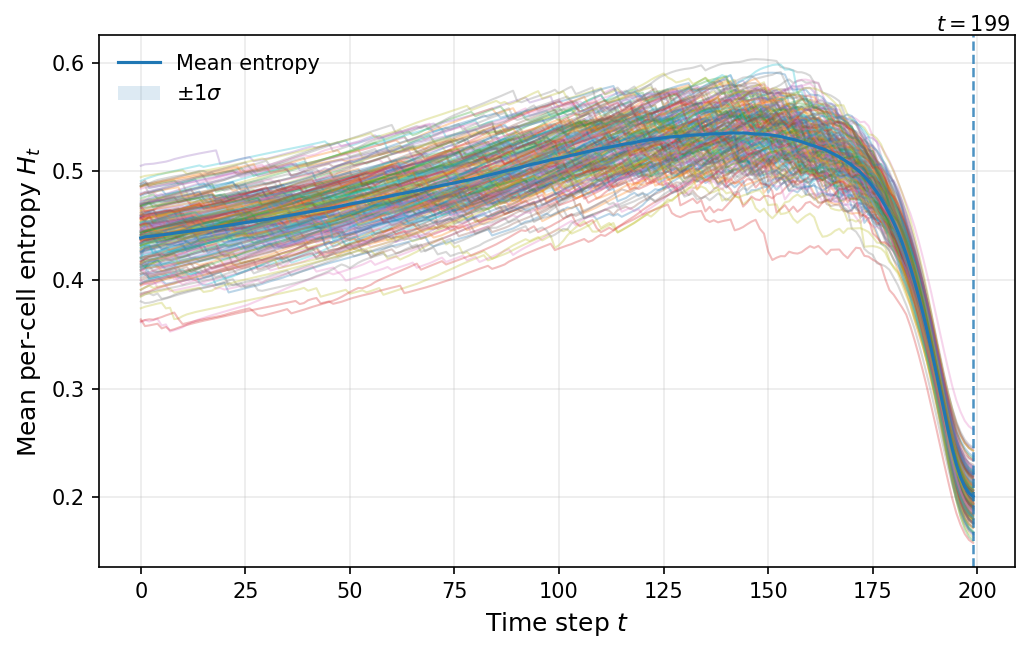} &
\includegraphics[width=0.30\textwidth]{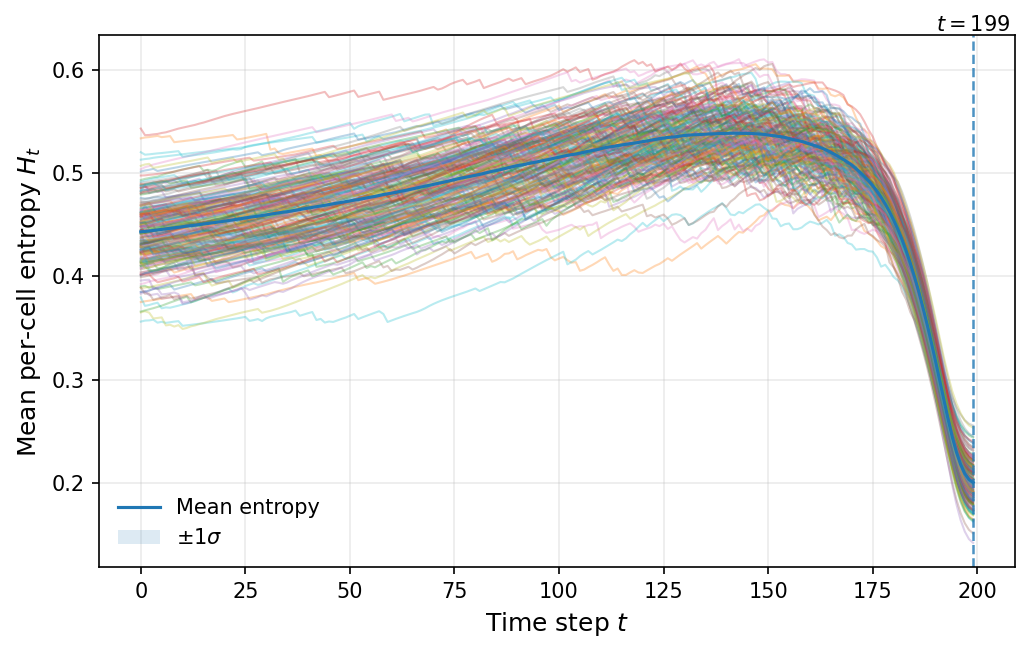} &
\includegraphics[width=0.30\textwidth]{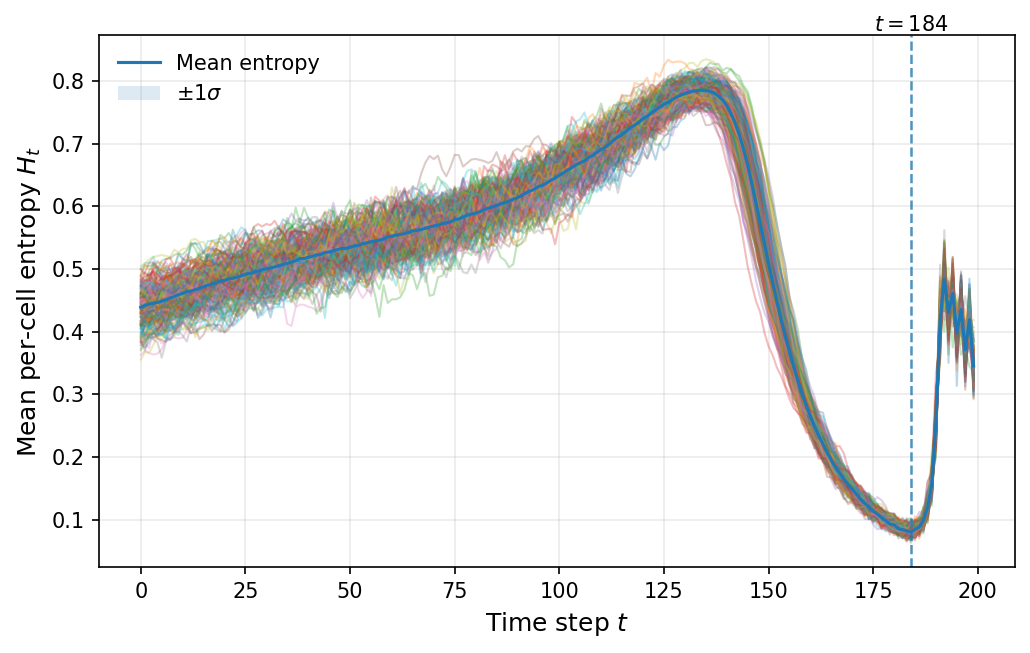} \\

\small ODE (score) &
\small DDPM sampler (DDIM) &
\small SDE (score+$\beta$) \\[6pt]

&
\includegraphics[width=0.30\textwidth]{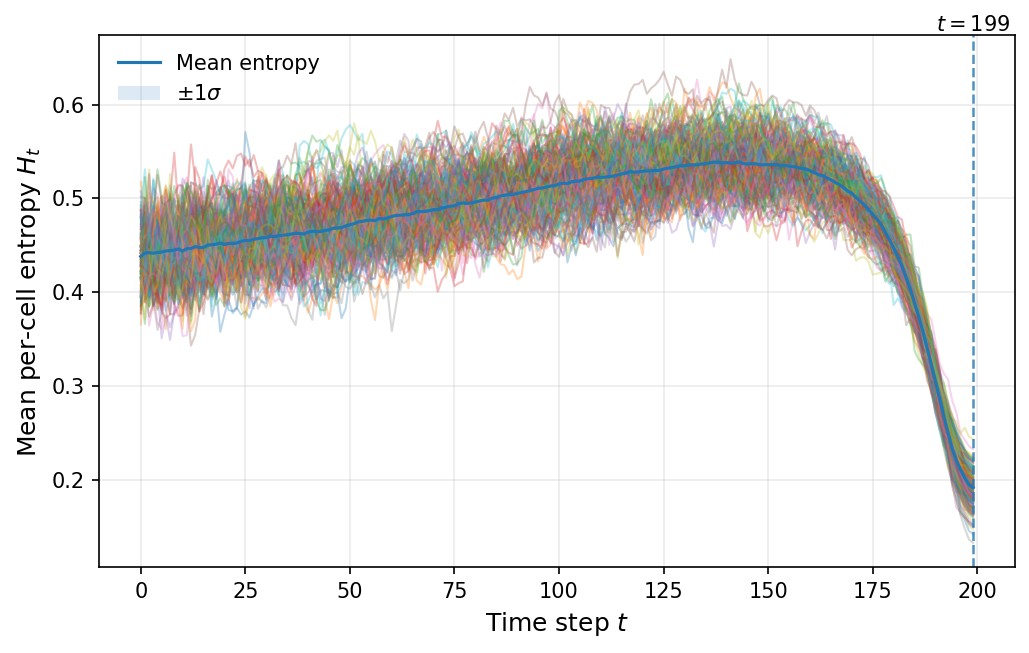} &
\includegraphics[width=0.30\textwidth]{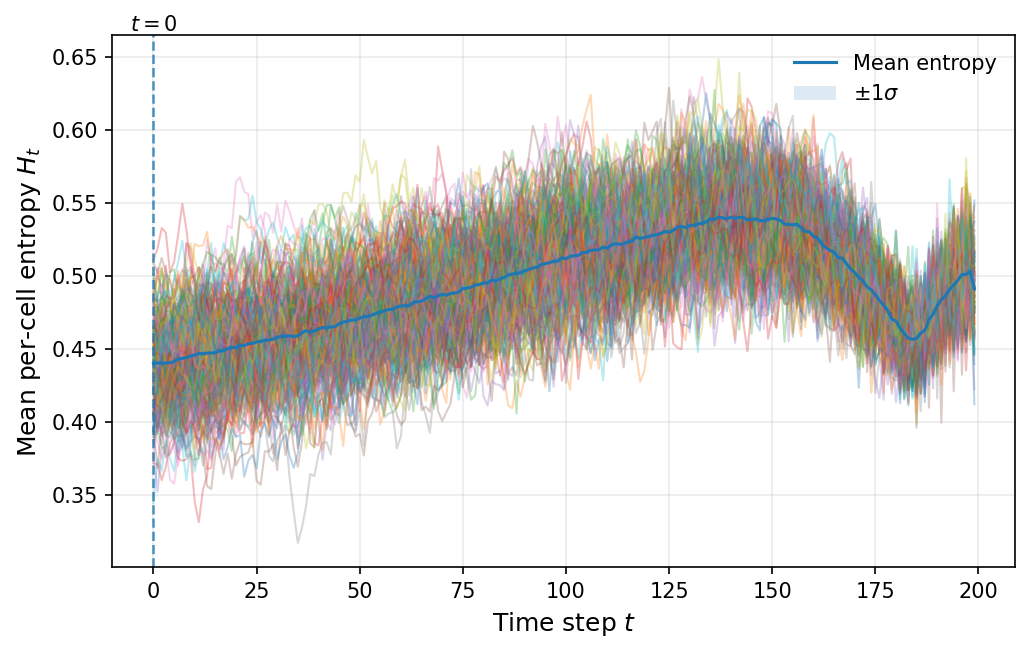} \\

&
\small DDPM sampler (DDPM) &
\small SDE (score)
\end{tabular}

\caption{
Entropy under cosine schedule for the Gaussian probability path.
Right column compares SDE score with diffusion coefficient equal to either $\sigma$ or $\beta(t)$.
Middle column compares DDPM samplers (DDIM vs DDPM).
ODE results (with a velocity field derived from the trained score) are shown adjacent to DDIM for reference.
}
\label{fig:diffusion_entropy_cosine}
\end{figure}

\paragraph{Cosine schedule for probability path coefficients and diffusion discretizations.}
Results under the cosine schedule are summarized in Table~\ref{tab:cosine_results} and
Figure~\ref{fig:diffusion_entropy_cosine}.
In this setting, we train a score model in continuous time along the cosine probability path
and evaluate both continuous-time samplers (ODE/SDE) and discrete-time diffusion samplers
(DDPM/DDIM) derived from the same model.

Among all unconditional generation methods considered, the cosine schedule combined with
a DDPM sampler achieves the highest success rate, exceeding $83\%$.
This represents a substantial improvement over all continuous-time ODE and SDE variants.

A plausible explanation for this gap lies in how sampling errors accumulate.
DDPM and DDIM samplers operate directly on the same discrete marginal distributions that the
model was trained to denoise, ensuring consistency between training and inference.
In contrast, continuous-time SDE solvers integrate an approximate drift field using numerical
methods; even small score estimation errors can accumulate over time, gradually drifting the
trajectory away from the data manifold.

We note that all diffusion-based results in this section are reported with dropout enabled
during evaluation.
Without dropout, the DDIM sampler collapses to a single stationary configuration
(see Appendix~\ref{app:ddim}), indicating that stochastic regularization remains important
even at inference time in this regime.

\subsection{Guided Sudoku Solving}

In addition to unconditional generation, we evaluate whether the learned generative models can be
repurposed as stochastic solvers for partially observed Sudoku puzzles.
In this guided setting, the objective is no longer to sample from the full data distribution,
but to generate any configuration consistent with a given set of fixed digits.
This provides a complementary test of whether the learned dynamics assign non-zero probability
mass to valid completions under hard global constraints.

\paragraph{Guided sampling setup.}
Given a partially filled Sudoku grid, we construct a binary mask indicating fixed (given)
cells and free variables.
Sampling is performed using guided SDE or diffusion-based samplers, where constraints are
enforced through masked updates during the simulation (we kept those as they performed the best).
We employ a two-stage constraint strategy controlled by a threshold
$\tau \in [0,1]$, which determines when constraints transition from
noise-consistent (soft) enforcement to exact clamping.
Early in the trajectory, fixed digits are enforced in a probabilistically consistent manner,
while at later stages they are imposed deterministically to ensure exact satisfaction.

\paragraph{Sampling protocol.}
For each Sudoku instance, we repeatedly sample batches of $512$ trajectories.
After each batch, all samples are discretized via per-cell argmax and checked for Sudoku
validity.
If at least one valid solution is found, sampling terminates early; otherwise, a new batch
is generated.
The threshold $\tau$ is sampled independently for each run from the set
$\{0.0, 0.45, 0.5, 0.55\}$, reflecting empirical observations that different puzzles and
models favor different transition points between soft and hard constraint enforcement.

\paragraph{Observed behavior.}

\begin{figure}[t]
    \centering
    \includegraphics[width=\linewidth]{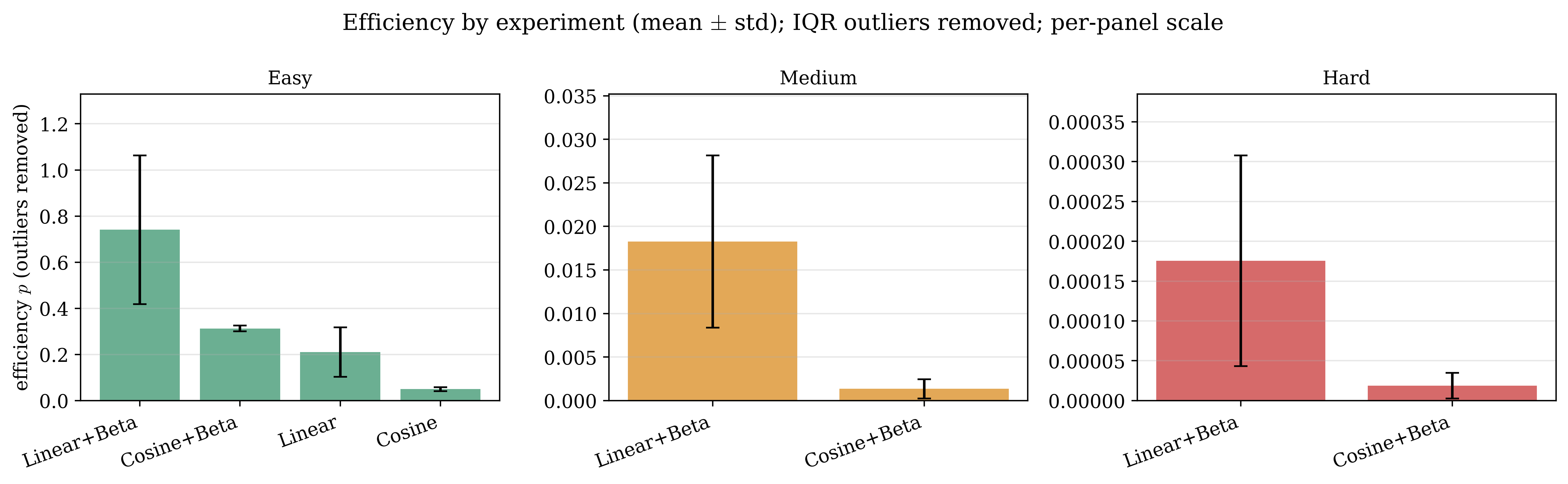}
    \caption{
    Efficiency comparison across noise schedules and diffusion parameterizations
    for Easy, Medium, and Hard Sudoku instances.
    Bars show mean efficiency over runs with one standard deviation.
    Linear schedules consistently outperform cosine schedules, motivating their use in
    subsequent experiments.
    }
    \label{fig:efficiency}
\end{figure}

Guided stochastic sampling consistently assigns
non-zero probability mass to valid completions even for difficult Sudoku instances.
As a result, repeated sampling eventually produces a correct solution for every puzzle
considered.
However, the number of sampling batches required varies substantially across instances:
some puzzles converge quickly, while others require many stochastic attempts and
correspondingly longer runtimes.
This variability motivates a closer examination of sampling \emph{efficiency}, rather than
mere success or failure.

\paragraph{Model and schedule choice.}
While cosine-scheduled SDEs performed strongly for unconditional generation, their behavior
in the guided solving setting is less predictable.
We therefore conduct a focused comparison using a score-based SDE model under two noise
schedules (linear and cosine) and two diffusion parameterizations:
a constant diffusion coefficient $\sigma$ and a time-dependent coefficient $\beta(t)$.

To quantify efficiency, we define a normalized success probability per instance,
\[
p \;=\; \frac{N_{\text{valid}}}{(r+1)\,N_{\text{batch}}},
\]
where $r$ is the number of failed sampling attempts prior to the first successful run,
$N_{\text{batch}}$ is the batch size (number of candidate grids evaluated per attempt),
and $N_{\text{valid}}$ denotes the number of valid Sudokus obtained in the first successful run.
All unsuccessful runs contribute zero valid solutions.
This metric jointly accounts for stochastic retries and batch evaluation cost, measuring the
fraction of valid solutions among all evaluated samples up to the first success.

Even on easy puzzles, this evaluation reveals a clear separation between methods.
Linear probability paths substantially outperform cosine schedules across both diffusion
choices.
In particular, the linear schedule combined with $\beta(t)$-scaled diffusion achieves
significantly higher efficiency than all other configurations
(Figure~\ref{fig:efficiency}).
Cosine-based variants, especially those using a constant diffusion coefficient, require
substantially more sampling and frequently fail to produce a valid solution within a
reasonable computational budget.

Based on these results, we restrict subsequent experiments on medium and hard puzzles to
the linear schedule with $\beta(t)$-scaled score-matching.
Other configurations are excluded, as they often require excessive computation and do not
reliably find solutions even after $1{,}000$ repetitions with a batch size of $512$.
The linear schedule continues to outperform alternative choices consistently as
puzzle difficulty increases.
Since DDPM-based samplers achieved the strongest performance in unconditional generation,
we compare the best-performing score-based guided solver against DDPM in the following
experiments.

\paragraph{Hard Sudoku instances and runtime.}

We compare the score-based SDE solver with the DDPM-based solver on hard Sudoku
instances, focusing on the computational effort required to obtain the first valid solution.
Both methods consistently succeed, but require multiple stochastic sampling runs.

\begin{figure}[t]
    \centering
    \begin{subfigure}{0.49\linewidth}
        \includegraphics[width=\linewidth]{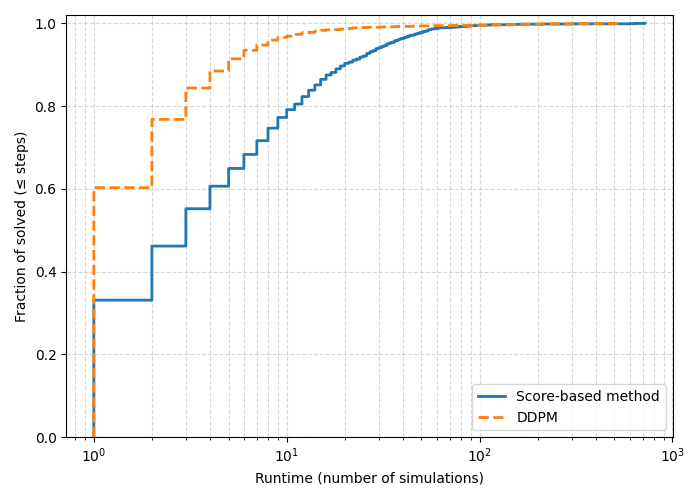}
        \caption{Runtime measured in number of sampling batches}
        \label{images:runtime_in_batches}
    \end{subfigure}
    \hfill
    \begin{subfigure}{0.49\linewidth}
        \includegraphics[width=\linewidth]{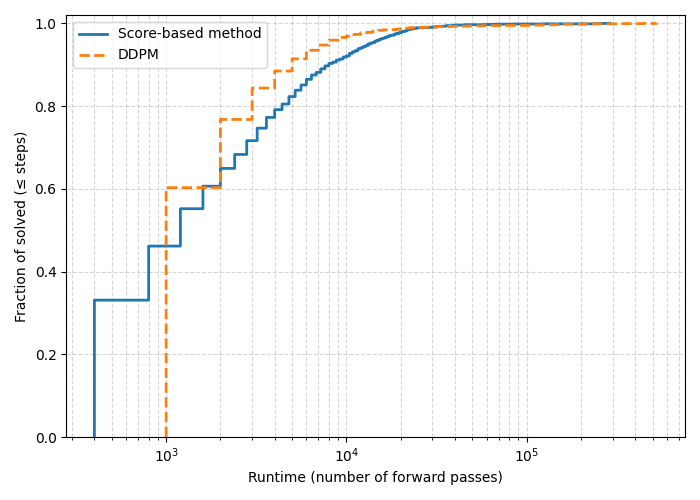}
        \caption{Runtime measured in total denoising steps}
        \label{images:runtime_in_passes}
    \end{subfigure}
    \caption{Empirical CDF of solving time on hard Sudoku instances under two runtime measures.}
\end{figure}

We evaluate on 2,280 hard Sudoku puzzles, running both methods with a batch size of 512
until a valid solution is found.
Because the two solvers operate with different denoising schedules (400 steps per trajectory
for the score-based solver and 1,000 steps for DDPM), we report runtime using two complementary
metrics.
First, we measure \emph{search efficiency} in terms of the number of sampling batches required
until a valid solution is found (Figure~\ref{images:runtime_in_batches}).
Second, we measure \emph{inference cost} as the total number of forward passes executed up to
the first success (Figure~\ref{images:runtime_in_passes}).

For the score-based solver with $\beta(t)$-scaled noise, we observe an average of 8.9 batches
before success, corresponding to approximately
$8.9 \times 512 \times 400 \approx 1.8 \times 10^6$ model evaluations per puzzle.
For DDPM, performance improves with longer trajectories; using 1,000 denoising steps per
trajectory, we observe an average of 3.5 batches before success, resulting in a comparable
average total of approximately $1.8 \times 10^6$ model evaluations.

Although the two methods have similar \emph{average} computational cost under their
respective best-performing configurations, their runtime distributions differ.
Under both runtime measures, the DDPM-based solver consistently dominates the score-based
solver, indicating that it reaches valid solutions in fewer sampling batches and, in most
cases, with fewer total denoising steps.

We emphasize that no early stopping is performed within a batch; all denoising trajectories are always executed in full. Thus, differences in runtime arise solely from the number of batches required.
In particular, DDPM solves a larger fraction of instances within any fixed step budget,
reflecting a higher concentration of probability mass on valid solutions.
Increasing the number of steps for the score-based SDE does not improve performance and
typically degrades results.

\section{Discussion}

We studied continuous-time generative models based on flow matching and score matching,
together with discrete diffusion models (DDPM/DDIM), using
Sudoku as a benchmark for globally constrained discrete structure.
Our experiments focused on unconditional generation and on guided Sudoku solving as a
test of whether learned generative dynamics can be repurposed for constraint satisfaction.

\paragraph{Unconditional generation.}
For Sudoku generation, stochasticity is essential.
Across probability paths and parameterizations, stochastic SDE sampling consistently
outperformed deterministic ODE sampler, highlighting the need for noise-driven
exploration to traverse the extremely sparse manifold of valid grids.
Score-based models were substantially more effective than flow-only models, suggesting
that explicitly learning the score  better captures
global constraints.

Among continuous-time methods, cosine probability paths performed best, and score-based
SDEs under this schedule achieved the highest success rates.
However, discrete DDPM and DDIM samplers significantly outperformed all continuous-time
variants, exceeding $80\%$ valid samples per batch.
This result indicates that, even for highly structured discrete data, diffusion models
operating directly on discrete-time marginal distributions remain extremely effective
generators.
We note that stable DDIM sampling required dropout at inference (with the same rate of $0.01$ as during training);
without dropout, sampling collapsed to a degenerate fixed point, with trajectories
converging to the same solution regardless of noise. Even small dropout of $0.01$ helps to avoid the collapse.

\paragraph{Guided Sudoku solving.}
The empirical behavior changes when applied to sudoku solving (guidance by given cells).
While cosine schedules are optimal for unconditional generation, they perform poorly
for guided Sudoku solving under our constraint-injection scheme.
Instead, linear probability paths combined with score-based SDE sampling and
$\beta(t)$-scaled noise yield more reliable solutions within reasonable time.

In this setting, DDPM-based solvers remain competitive.
Although DDPM requires longer denoising trajectories, it compensates by requiring fewer
independent sampling attempts.
As a result, the two approaches exhibit comparable average inference cost under their
best-performing configurations, despite very different noise schedules and dynamics.
Notably, the large performance gap observed in unconditional generation (e.g.\ DDPM
versus linear SDEs) largely disappears once hard constraints are imposed.
This suggests that conditioning fundamentally alters the effective search dynamics and
reduces the advantage of marginal-consistent diffusion discretizations.
Understanding why samplers that do not strictly follow the probability path can nevertheless
be competitive for constraint satisfaction is an interesting open research question.

\paragraph{Interpretation.}
The guided solver does not perform symbolic reasoning or explicit constraint propagation.
Instead, it repeatedly samples from a learned continuous relaxation of the Sudoku solution
space, relying on stochastic exploration.

From an algorithmic perspective, this approach is inefficient compared to classical
Sudoku solvers.
However, it demonstrates that generative models trained purely on completed solutions can
be repurposed as stochastic constraint solvers without task-specific heuristics or search
procedures.
The contrast between unconditional generation and guided solving suggests that different
diffusion parameterizations may favor either faithful sampling of the data distribution
or effective exploration under hard constraints.

\paragraph{Limitations and future work.}
This study is exploratory.
We did not perform systematic ablations of the guided constraint mechanism, nor did we
conduct a comprehensive comparison against specialized representations designed for
discrete reasoning (e.g. \cite{avdeyev2023dirichlet}). Our goal is to isolate the behavior of standard diffusion and flow formulations when applied without specialized representations.
Experiments were limited to a single dataset and model architecture, and results may not
generalize directly to other combinatorial domains.

Future work could explore principled methods for constraint injection, adaptive noise
control, and hybrid approaches that combine generative sampling with symbolic reasoning
or search.
A deeper theoretical understanding of why diffusion dynamics that deviate from the
probability path can outperform path-consistent dynamics under conditioning remains an
important open direction.

\section{Conclusion}

We studied continuous-time generative models based on flow and score matching,
together with discrete diffusion models, on Sudoku as a benchmark for highly
constrained discrete structure.
Our results show that, despite the extreme sparsity of valid solutions,
diffusion-based models can assign meaningful probability mass to globally
consistent configurations.

For unconditional generation, stochastic sampling is essential and discrete
diffusion models (DDPM/DDIM) remain the most effective generators, substantially
outperforming continuous-time SDEs.
In contrast, under hard conditioning, the performance gap between methods
narrows: score-based SDEs with simple linear schedules and annealed noise become
competitive with DDPM despite such SDEs deviating from probability-path-consistent
dynamics.

These findings suggest that different diffusion parameterizations favor either
faithful sampling of the data distribution or effective stochastic exploration
under constraints.
Sudoku provides a compact but demanding testbed for revealing this distinction,
highlighting both the promise and current limitations of diffusion-based
generative modeling for structured reasoning tasks.

\bibliographystyle{plain} 
\bibliography{output}

\appendix 

\section{Model}
\label{app:model}

\paragraph{Time embedding.}
We condition the network on the continuous time variable $t\in[0,1]$ using Fourier features.
Concretely, we map $t$ to a $d_t=64$ dimensional embedding using sinusoidal components at exponentially spaced frequencies up to $f_{\max}=10^3$:
\[
\phi(t)=\big[\sin(2\pi f_1 t),\dots,\sin(2\pi f_{d_t/2} t),\; \cos(2\pi f_1 t),\dots,\cos(2\pi f_{d_t/2} t)\big],
\]
where $f_k=\exp\!\big(\mathrm{linspace}(0,\log f_{\max}, d_t/2)_k\big)$.
The Fourier features are then passed through a two-layer MLP,
\[
\mathrm{MLP}_t:\mathbb{R}^{d_t}\to\mathbb{R}^{H},\qquad
\mathrm{MLP}_t(x)=W_2\,\mathrm{SiLU}(W_1 x),
\]
with hidden dimension $H=128$.
The resulting time-conditioning vector in $\mathbb{R}^{H}$ is added to every token embedding.

\paragraph{Transformer backbone.}
The backbone consists of $L=4$ standard Transformer blocks with hidden dimension $H=128$ and $8$ attention heads.
Each block applies LayerNorm, multi-head self-attention (with dropout $0.01$), and a position-wise MLP with GELU nonlinearity and expansion ratio $4$ (i.e., intermediate width $4H$), followed by residual connections.
Dropout ($0.01$) is applied to the attention and MLP residual branches.
This architecture enables global interactions between cells, which is essential for Sudoku due to its non-local row/column/box constraints.

\paragraph{Sudoku-aware positional encoding.}
Unlike generic sequence tasks, Sudoku tokens have a known 2D structure and a third type of locality induced by the $3\times 3$ subgrids.
We therefore use three learned embedding tables for row index, column index, and box index, each mapping $\{0,\dots,8\}$ to $\mathbb{R}^{H}$.
For a cell at position $(r,c)$ with box index $b(r,c)$, the positional vector is
\[
e_{\mathrm{pos}}(r,c)=e_{\mathrm{row}}(r)+e_{\mathrm{col}}(c)+e_{\mathrm{box}}(b(r,c)),
\]
which is added to the projected token embedding.
This provides the model explicit access to Sudoku geometry and biases attention toward respecting row/column/box structure.

\section{Deterministic DDIM and Dropout}
\label{app:ddim}

In all other experiments in this work, dropout was kept enabled at inference time. Here, we analyze the effect of disabling dropout in DDPM-like samplers, and show that doing so reveals a qualitative failure mode for deterministic solvers.

All samplers use the same trained score network and the same time discretization. For DDIM, we consider two inference modes:
(i) \emph{deterministic}, where the noise term is set to zero at each reverse step and dropout is disabled, and
(ii) \emph{stochastic}, where dropout remains enabled during inference, introducing stochasticity through the network. DDPM sampling remains stochastic by construction. All methods are evaluated on unconditional Sudoku generation with 25\,000 samples.

Table~\ref{tab:ddim_ddpm_dropout} summarizes validity and diversity statistics. Validity is measured as the fraction of generated boards satisfying all Sudoku constraints, and diversity is measured by the number of unique valid solutions.

\begin{table}[t]
\centering
\begin{tabular}{lcccc}
\toprule
Sampler & Dropout & Total samples & Valid samples & Unique solutions \\
\midrule
DDPM & yes & 25\,000 & 20\,962 & 20\,962 \\
DDPM & no  & 25\,000 & 21\,350 & 21\,350 \\
DDIM & yes & 25\,000 & 19\,926 & 19\,926 \\
DDIM & no  & 25\,000 & 25\,000 & 1 \\
\bottomrule
\end{tabular}
\caption{Comparison of DDPM and DDIM sampling with and without dropout at inference. Deterministic DDIM achieves perfect validity but collapses to a single solution.}
\label{tab:ddim_ddpm_dropout}
\end{table}

\begin{figure}[t]
    \centering
    \includegraphics[width=0.3\linewidth]{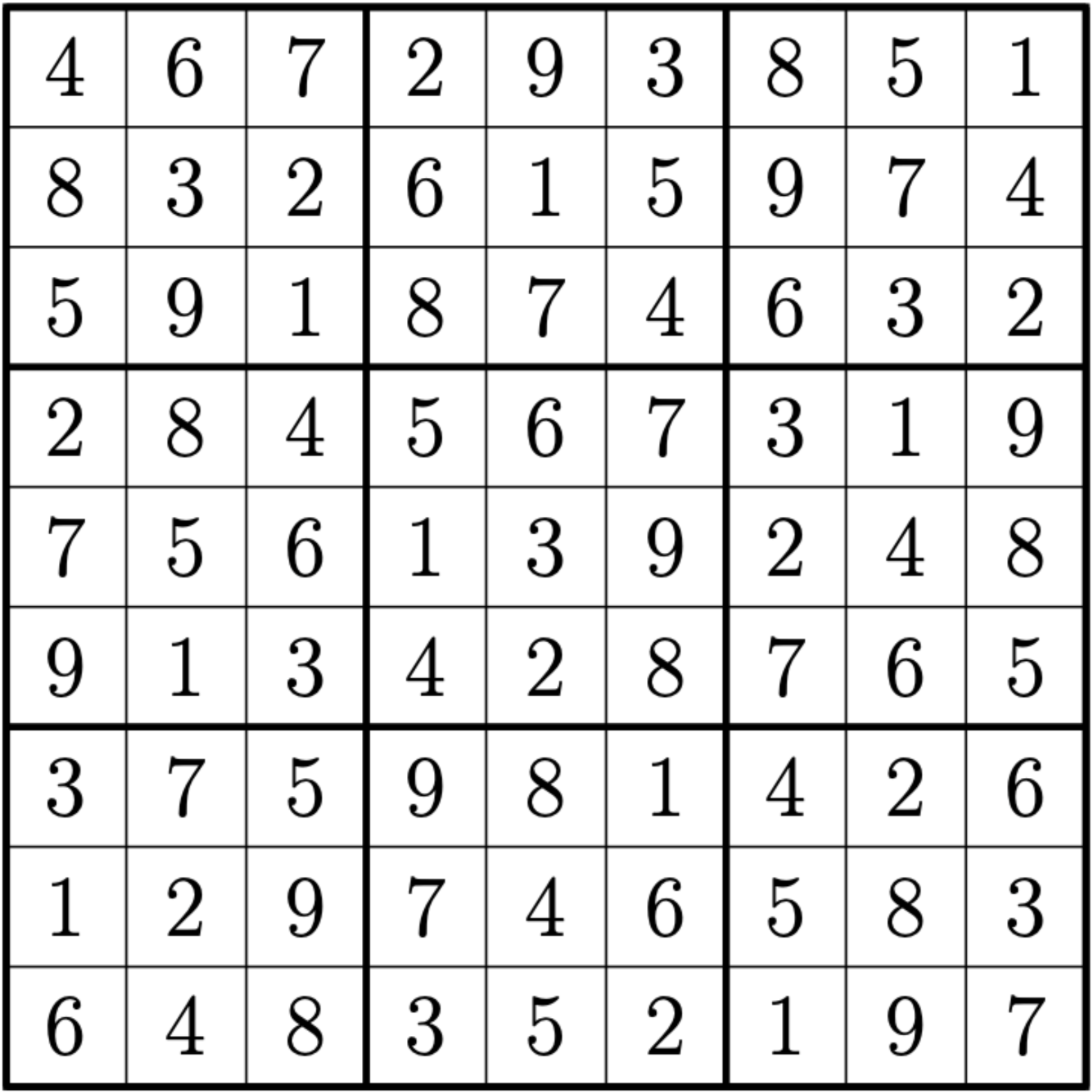}
    \caption{
    Example of a stationary Sudoku solution produced by deterministic DDIM. All samples converge to the same valid grid regardless of the initial noise.
    }
    \label{fig:solving_hard}
\end{figure}

When dropout is disabled, DDIM produces a single valid Sudoku solution across all samples, despite different initial noise realizations. This behavior indicates convergence to a stationary attractor of the deterministic reverse process, rather than sampling from the full distribution of valid solutions. An example of such a stationary solution is shown in Figure~\ref{fig:solving_hard}.

\section{Theoretical groundings}

\subsection{Flow matching}

We adopt a Gaussian probability path that interpolates between the data distribution and a standard Gaussian. For a data point $z \in \mathbb{R}^d$, we consider a conditional probability path of the form
\[
p_t(x \mid z) = \mathcal{N}(x; \alpha_t z, \beta_t^2 I),
\]
where $\alpha_t$ and $\beta_t$ are continuous functions satisfying the boundary conditions $\alpha_0 = 0$, $\beta_0 = 1$, $\alpha_1 = 1$, and $\beta_1 = 0$. Sampling from this path can be written explicitly as
\begin{equation}
x_t = \alpha_t z + \beta_t \varepsilon, \qquad \varepsilon \sim \mathcal{N}(0, I),
\label{eq:app:path}
\end{equation}

where $x_{1}$ corresponds to data.

A central result in flow-based generative modeling is that, given a probability path $p_t$, one can construct a corresponding vector field whose induced ordinary differential equation transports samples along this path. In particular, for the conditional Gaussian probability path defined above, there exists a conditional target velocity field $u_t^{\mathrm{target}}(x \mid z)$ such that solutions of
\[
\frac{d x_t}{d t} = u_t^{\mathrm{target}}(x_t \mid z)
\]
satisfy $x_t \sim p_t(\cdot \mid z)$ when initialized from $x_0 \sim \mathcal{N}(0, I)$.

For Gaussian probability paths, this conditional velocity field admits a closed-form expression,
\[
u_t^{\mathrm{target}}(x \mid z)
=
\left( \dot{\alpha}_t - \frac{\dot{\beta}_t}{\beta_t} \alpha_t \right) z
+
\frac{\dot{\beta}_t}{\beta_t} x,
\]
where dots denote derivatives with respect to time. While the marginal velocity field obtained by integrating over the data distribution is generally intractable, an important result from flow matching theory shows that minimizing a regression loss against the conditional velocity field differs from minimizing the marginal loss only by a constant. As a consequence, learning to approximate the conditional velocity field is sufficient to recover the correct marginal dynamics.

In practice, we parameterize a neural network $u_\theta(x,t)$ and train it to approximate the conditional target velocity field under the Gaussian probability path. Once trained, generation proceeds by simulating the flow defined by the ordinary differential equation
\[
\frac{d x_t}{d t} = u_\theta(x_t, t),
\]
starting from Gaussian noise at $t=0$ and integrating forward to $t=1$. This constitutes the flow matching component of our approach and serves as the foundation for our experiments on unconditional Sudoku generation.

\subsection{Score Matching and Diffusion Models}

In addition to flow matching, the same probability path can be modeled using stochastic differential equations. Rather than describing the evolution of samples via a deterministic ordinary differential equation, diffusion-based models introduce stochasticity by adding a Brownian motion term. Concretely, given a probability path $p_t$, one considers stochastic dynamics of the form
\[
dX_t
=
\left[
u_t^{\mathrm{target}}(X_t)
+
\frac{\sigma_t^2}{2} \nabla \log p_t(X_t)
\right] dt
+
\sigma_t \, dW_t,
\]
where $\sigma_t$ is a time-dependent diffusion coefficient and $W_t$ denotes a standard Brownian motion. From theory, it follows that this stochastic differential equation has marginal distributions $X_t \sim p_t$ for all $t \in [0,1]$, provided the drift term is chosen as above.

The term $\nabla \log p_t(x)$ is known as the \emph{score function}. While the marginal score function $\nabla \log p_t(x)$ is generally intractable, it can be expressed as an expectation over conditional score functions. As in the flow matching case, minimizing a loss against the conditional score function is sufficient, as the corresponding marginal objective differs only by an additive constant.

For the Gaussian probability path
\[
p_t(x \mid z) = \mathcal{N}(x; \alpha_t z, \beta_t^2 I),
\]
the conditional score function admits a closed-form expression,
\[
\nabla \log p_t(x \mid z)
=
- \frac{x - \alpha_t z}{\beta_t^2}.
\]
This expression can be used directly as a training target for a neural network approximating the score function.

\subsubsection{Connection to Denoising Diffusion Models}

\paragraph{Langevin dynamics (static probability path).}
The \cite{holderrieth2026flow}\cite{holderrieth2025introductionflowmatchingdiffusion} introduce an ``SDE extension'' of a probability path $p_t$ via the Fokker--Planck equation. In particular, for an It\^{o} SDE
\[
dX_t = b_t(X_t)\,dt + \sigma_t\,dW_t,
\]
the associated density $p_t$ satisfies the Fokker--Planck equation
\[
\partial_t p_t(x) = -\nabla\cdot\!\big(p_t(x)\,b_t(x)\big) + \frac{\sigma_t^2}{2}\Delta p_t(x).
\]
The notes then show that for a (possibly time-dependent) path $p_t$, choosing the drift as
\[
b_t(x)=u_t^{\mathrm{target}}(x) + \frac{\sigma_t^2}{2}\nabla\log p_t(x)
\]
yields an SDE whose marginals follow $p_t$. As a special case (Remark~16 in \cite{holderrieth2026flow}\cite{holderrieth2025introductionflowmatchingdiffusion}), if the path is \emph{static}, i.e.\ $p_t(x)=p(x)$ for all $t$ (equivalently $\partial_t p_t(x)=0$), one may set $u_t^{\mathrm{target}}\equiv 0$ and obtain
\[
dX_t = \frac{\sigma_t^2}{2}\nabla\log p(X_t)\,dt + \sigma_t\,dW_t,
\]
which the notes identify as \emph{Langevin dynamics}. In this static setting, $p$ is a stationary distribution of the dynamics: if $X_0\sim p$, then $X_t\sim p$ for all $t\ge 0$, and under suitable regularity conditions the dynamics converge to $p$ from other initializations. Importantly, in the static Langevin setting $\sigma_t$ can be chosen freely (it affects mixing speed rather than the stationary distribution).

\paragraph{Why static Langevin is not sufficient for our time-varying probability path.}
In our experiments we work with a time-varying Gaussian probability path $p_t$ (Section~1), where $p_t$ changes with $t$ (e.g.\ through $\alpha_t$ and $\beta_t$). The static Langevin special case does not apply to such paths. In fact, if one attempts to remove the transport term while keeping a time-dependent target, i.e.\ sets
\[
b_t(x)=\frac{\sigma_t^2}{2}\nabla\log p_t(x)
\qquad\text{with}\qquad \sigma_t=\sigma(t),
\]
then the Fokker--Planck equation forces the path to be static: using the identity
$p_t(x)\nabla\log p_t(x)=\nabla p_t(x)$, one obtains
\[
-\nabla\cdot\!\Big(p_t \frac{\sigma_t^2}{2}\nabla\log p_t\Big)+\frac{\sigma_t^2}{2}\Delta p_t
=
-\frac{\sigma_t^2}{2}\Delta p_t + \frac{\sigma_t^2}{2}\Delta p_t
=0,
\]
hence $\partial_t p_t(x)=0$. Therefore, \emph{no choice of a purely time-dependent diffusion coefficient $\sigma_t$ can compensate for the absence of transport when the desired marginals $p_t$ vary with $t$}. This clarifies why the Langevin dynamics from \cite{holderrieth2026flow}\cite{holderrieth2025introductionflowmatchingdiffusion} should be interpreted as a stationary sampler for a fixed distribution rather than as a finite-time noise-to-data bridge.

\section{Samplers}
\label{app:samplers}

This appendix provides explicit discrete-time update equations for the sampling procedures
described in Section~\ref{subsec:sampling}.

\paragraph{ODE sampling (Euler).}
\[
x_{t_{k+1}} = x_{t_k} + h_k\,u_\theta(x_{t_k},t_k).
\]

where $u_\theta(x_{t_k},t_k)$ is the learned flow field.

\paragraph{SDE sampling (Euler--Maruyama).}

\[
x_{t_{k+1}}
=
x_{t_k}
+
h_k
\Big(
u_\theta(x_{t_k},t_k)
+
\frac{1}{2}\,\sigma^2\,s_\theta(x_{t_k},t_k)
\Big)
+
\sigma\sqrt{h_k}\,\varepsilon_k,
\qquad
\varepsilon_k\sim\mathcal N(0,I).
\]
where $s_\theta(x,t)\approx\nabla_x\log p_t(x)$ is a learned score function and
$\sigma>0$ is a constant diffusion coefficient.

\paragraph{Empirical $\beta(t)$-diffusion sampler.}

\[
x_{t_{k+1}}
=
x_{t_k}
+
h_k
\Big(
u_\theta(x_{t_k},t_k)
+
\frac{1}{2}\,\sigma^2\,s_\theta(x_{t_k},t_k)
\Big)
+
\beta(t_k)\sqrt{h_k}\,\varepsilon_k,
\qquad
\varepsilon_k\sim\mathcal N(0,I).
\]

\section{DDPM-style discretization from a continuous Gaussian path}
\label{app:ddpm_from_path}

This appendix details how a DDPM-style ancestral sampler can be derived from the continuous
Gaussian probability path used for training.

We train a score model
\begin{equation}
s_\theta(x,t) \approx \nabla_x \log p_t(x),
\label{eq:score_def}
\end{equation}
along the continuous path \eqref{eq:app:path}. For Gaussian perturbations, the conditional score is
\begin{equation}
\nabla_x \log p_t(x\mid x_0) = \frac{\alpha(t)x_0 - x}{\beta(t)^2}.
\label{eq:cond_score}
\end{equation}
Rearranging \eqref{eq:path} gives $x-\alpha(t)x_0=\beta(t)\varepsilon$, hence
\[
\nabla_x \log p_t(x\mid x_0) = -\frac{\varepsilon}{\beta(t)}.
\]
This motivates converting a learned score into a noise predictor,
\begin{equation}
\varepsilon_\theta(x,t) = -\beta(t)\, s_\theta(x,t).
\label{eq:eps_from_score}
\end{equation}

To obtain a discrete sampler, we evaluate the continuous-time model on a grid
$\{t_k\}_{k=0}^T$ with $t_0=0$ and $t_T=1$, and define
\begin{equation}
\bar\alpha_k := \alpha(t_k)^2,
\qquad
1-\bar\alpha_k := \beta(t_k)^2.
\label{eq:discrete_bar_alpha}
\end{equation}
Then the corresponding forward marginals satisfy
\[
q(x_k\mid x_0) = \mathcal{N}\!\big(\sqrt{\bar\alpha_k}\,x_0,\;(1-\bar\alpha_k)I\big),
\]
matching the standard DDPM form.

We use an \emph{ancestral} update that inverts the marginal noising process.
Given $x_{t_k}$ we compute an estimate of the clean sample,
\[
\hat{x}_1
=
\frac{x_{t_k} - \beta(t_k)\,\varepsilon_\theta(x_{t_k},t_k)}{\alpha(t_k)}.
\]
We then resample at the next time point using the same Gaussian interpolation law:
\[
x_{t_{k+1}}
=
\alpha(t_{k+1})\,\hat{x}_1
+
\beta(t_{k+1})\,\varepsilon,
\qquad \varepsilon\sim\mathcal{N}(0,I).
\]
This sampler is aligned with the continuous-time training objective and is less prone to the
error accumulation observed when directly simulating a discrete Markovian reverse chain
(Appendix~\ref{app:chain}).

\section{DDPM from continuous score models}
\label{app:ddpm}

\paragraph{From $\bar\alpha_k$ to $(\alpha_k,\beta_k)$.}
In Appendix~\ref{app:ddpm_from_path} we work directly with the forward \emph{marginals}
$q(x_k\mid x_0)=\mathcal N(\sqrt{\bar\alpha_k}\,x_0,(1-\bar\alpha_k)I)$ induced by the
continuous Gaussian path at discrete times $t_k$.
Classical DDPMs instead define a \emph{Markov} forward process with per-step transitions
\[
q(x_k\mid x_{k-1})=\mathcal N\!\big(\sqrt{\alpha_k}\,x_{k-1},\;\beta_k I\big),
\qquad \beta_k = 1-\alpha_k,
\]
with independent Gaussian noise increments.
Unrolling the recursion yields
\[
x_k
=
\sqrt{\alpha_k}x_{k-1} + \sqrt{\beta_k}\,\varepsilon_k
=
\Big(\prod_{i=1}^k \sqrt{\alpha_i}\Big)x_0 + \text{Gaussian noise},
\]
so the coefficient multiplying $x_0$ in $q(x_k\mid x_0)$ equals $\prod_{i=1}^k\sqrt{\alpha_i}$.
Matching this with the marginal form implies
\[
\sqrt{\bar\alpha_k}=\prod_{i=1}^k\sqrt{\alpha_i}
\quad\Longrightarrow\quad
\bar\alpha_k=\prod_{i=1}^k \alpha_i.
\]
Therefore, given a sequence $\{\bar\alpha_k\}_{k=0}^T$ one can recover the Markov
coefficients via
\begin{equation}
\alpha_k=\frac{\bar\alpha_k}{\bar\alpha_{k-1}},
\qquad
\beta_k = 1-\alpha_k.
\label{eq:ak_bk}
\end{equation}
In our experiments, we refer to samplers that explicitly simulate a discrete reverse chain
using these per-step parameters as \emph{Markovian} DDPM/DDIM samplers (Appendix~\ref{app:chain}).
By contrast, the \emph{ancestral} sampler in Appendix~\ref{app:ddpm_from_path} directly inverts
the forward marginals and does not rely on the Markov factorization.

\paragraph{Diffusion-compatible (trigonometric) schedules.}
A necessary condition for a continuous Gaussian path $x_t=\alpha(t)x_0+\beta(t)\varepsilon$
to match diffusion marginals at discrete times $t_k$ is
\[
\alpha(t)^2+\beta(t)^2=1 \quad \text{for all } t,
\]
since DDPM marginals satisfy $\mathrm{Var}(x_t\mid x_0)=1-\bar\alpha_t$ with
$\bar\alpha_t=\alpha(t)^2$.
The linear choice $\alpha(t)=t$, $\beta(t)=1-t$ violates this condition and therefore does not
correspond to any DDPM forward process (although it defines a valid Gaussian interpolation).
A diffusion-compatible alternative is obtained by parameterizing $(\alpha,\beta)$ on the unit circle, e.g.
\[
\alpha(t)=\sin\!\Big(\frac{\pi}{2}t\Big),
\qquad
\beta(t)=\cos\!\Big(\frac{\pi}{2}t\Big),
\]
which satisfies $\alpha(t)^2+\beta(t)^2=1$ identically and induces $\bar\alpha_k=\alpha(t_k)^2$.
In our implementation we index the reverse-time sampler from noise to data; this is equivalent to applying
$t\leftarrow 1-t$ in the schedule and only affects indexing conventions.

\section{Why the direct DDPM chain performs poorly}
\label{app:chain}
We observed that simulating a Markovian reverse chain directly from the score 
function leads to very poor results (around $0.024$ valid Sudokus per $1$ generation batch). 
Surprisingly, multiplying the predicted score or noise by a global factor 
(e.g.\ $\varepsilon_\theta \mapsto 8\varepsilon_\theta$ or even $15\varepsilon_\theta$) 
dramatically improves the success rate (up to $0.75$ valid Sudokus per batch). 
This initially appeared paradoxical, because the DDPM reverse recurrence is 
mathematically correct under its assumptions; however, the behaviour turned out 
to be a classical and well-documented calibration mismatch between marginal 
training and Markov-chain simulation.

In principle, the DDPM reverse update
\[
x_{k-1}
=
\frac{1}{\sqrt{\alpha_k}}
\left(
x_k
-
\frac{1-\alpha_k}{\sqrt{1-\bar\alpha_k}}
\,\varepsilon_\theta(x_k,k)
\right)
+
\sigma_k \varepsilon,
\]
is valid only if the learned noise prediction 
$\varepsilon_\theta(x_k,k)$ is \emph{perfectly calibrated} with respect to the 
per-step forward diffusion noise. 
That is, the model must match both the \emph{direction} and the \emph{magnitude} 
of the Gaussian noise increment prescribed by the Markov chain.  
In practice, however, diffusion models rarely achieve this ideal calibration:  
the predicted noise magnitude is typically under-estimated, over-estimated, or 
varies with the timestep~$k$.
Since the DDPM chain applies these predictions iteratively, even a small bias in 
$\varepsilon_\theta$ compounds over the $T$ reverse steps, drifting the 
trajectory away from the data manifold and producing invalid samples.

\paragraph{Why this issue does not appear in SDE-based samplers.}
The same failure mode does \emph{not} occur when using SDE or ODE samplers.  
This is because SDE solvers operate directly on the \emph{continuous} 
probability path and only require the score at the current time $t$ as
they do not rely on the specific discrete Markovian factorization
\(
q(x_k|x_{k-1})
\).
Instead, SDE/ODE samplers integrate a differential equation driven by the score, 
which depends only on the instantaneous marginal 
distribution~$q(x_t)$ that the model was trained to match.
Thus, marginal-based training aligns naturally with continuous-time sampling, 
and calibration errors do not accumulate catastrophically as they do in the 
discrete DDPM chain.

\begin{table}[ht]
\centering
\caption{Comparison of DDPM/DDIM samplers.}
\label{tab:ddpm_ddim_markovian}

\scriptsize
\setlength{\tabcolsep}{-0.8pt}
\renewcommand{\arraystretch}{1.15}

\begin{tabular}{l *{8}{>{\centering\arraybackslash}p{1.35cm}}}
\hline
\textbf{Metric} &
\rotatebox{80}{DDPM, ancestral} &
\rotatebox{80}{DDIM, ancestral} &
\rotatebox{80}{DDIM, markovian} &
\rotatebox{80}{DDPM, markovian} &
\rotatebox{80}{DDPM, ancestral} &
\rotatebox{80}{DDIM, ancestral} &
\rotatebox{80}{DDIM, markovian} &
\rotatebox{80}{DDPM, markovian} \\
\hline
\hline

Rate (50 runs) &
0.8364 &
0.7880 &
0.0036 &
0.0265 &
0.5785 &
0.3572 &
0.0056 &
0.0312 \\

Std. of rate &
0.0138 &
0.0147 &
0.0027 &
0.0059 &
0.0258 &
0.0213 &
0.0020 &
0.0041 \\
\hline

Pearson corr &
-0.15 &
-0.89 &
-0.34 &
-0.27 &
-0.81 &
-0.96 &
-0.2260 &
-0.1888 \\

$p$-value &
$\approx 10^{-3}$ &
$\approx 10^{-174}$ &
$\approx 10^{-15}$ &
$\approx 10^{-9}$ &
$\approx 10^{-116}$ &
$\approx 10^{-266}$ &
$\approx 10^{-7}$ &
$\approx 10^{-5}$ \\
\hline

Steps &
400 &
400 &
400 &
400 &
200 &
200 &
200 &
200 \\
\hline
\end{tabular}
\end{table}

\paragraph{Markovian vs ancestral samplings comparison.}

We contrast two sampling strategies. 
\emph{Ancestral sampling} directly inverts the marginal noising process via
\[
\hat{x}_1 = \frac{x_t - \beta(t)\,\varepsilon_\theta(x_t,t)}{\alpha(t)},
\qquad
x_{t'} = \alpha(t')\,\hat{x}_1 + \beta(t')\,\varepsilon,
\]
where each step resamples from the same Gaussian interpolation model used in the
forward process.
By contrast, \emph{Markovian sampling} attempts to follow the discrete DDPM reverse
chain using the posterior update
\[
x_{k+1}
=
\frac{1}{\sqrt{\alpha_k}}
\!\left(
x_k - \frac{1-\alpha_k}{\sqrt{1-\bar{\alpha}_k}}
\,\varepsilon_\theta(x_k,k)
\right)
+ \sigma_k \varepsilon,
\]
which requires accurate calibration of per-step diffusion parameters.

In the Table \ref{tab:ddpm_ddim_markovian} compares DDPM and DDIM samplers under two regimes: ancestral sampling, which directly inverts the marginal noising process, and Markovian sampling, which attempts to simulate the discrete DDPM reverse chain. The results demonstrate that Markovian DDPM/DDIM sampling performs extremely poorly, with success rates near zero and weak correlations. In contrast, ancestral samplers perform dramatically better, reaching 0.83–0.79 validity at 400 steps and remaining significantly stronger even at 200 steps. Increasing the number of steps from 200 to 400 consistently improves performance for both DDPM and DDIM in the ancestral setting, indicating that finer temporal resolution stabilizes the reverse dynamics. Overall, the table shows that consecutive Markovian updates are not reliable in this setting, and that ancestral (marginal-consistent) sampling is essential for high-quality generation.

\end{document}